\documentclass[conference]{IEEEtran}
\IEEEoverridecommandlockouts
\usepackage{cite}
\usepackage{amsmath,amssymb,amsfonts}
\usepackage{algorithmic}
\usepackage{graphicx}
\usepackage{textcomp}
\usepackage{xcolor}
\usepackage{url}
\def\BibTeX{{\rm B\kern-.05em{\sc i\kern-.025em b}\kern-.08em
    T\kern-.1667em\lower.7ex\hbox{E}\kern-.125emX}}
\begin{document}

\title{Applications of Artificial Intelligence in Live Action Role-Playing Games (LARP)}

\author
{\IEEEauthorblockN{Christoph Salge}
\IEEEauthorblockA{\small\textit{University of Hertfordshire}\\
Hatfield, UK \\
c.salge@herts.ac.uk}
\and
\IEEEauthorblockN{Emily Short}
\IEEEauthorblockA{\small\textit{Failbetter Games}\\
Oxford, UK \\
emshortif@gmail.com}
\and
\IEEEauthorblockN{Mike Preuss}
\IEEEauthorblockA{\small\textit{Universiteit Leiden}\\
Leiden, The Netherlands \\
m.preuss@liacs.leidenuniv.nl}
\and
\IEEEauthorblockN{Spyridon Samothrakis}
\IEEEauthorblockA{\small\textit{University of Essex}\\
Colchester, UK \\
ssamot@essex.ac.uk}
\and
\IEEEauthorblockN{Pieter Spronck}
\IEEEauthorblockA{\small\textit{Tilburg University} \\
Tilburg, The Netherlands \\
p.spronck@uvt.nl}
}

\maketitle
\begin{abstract}

Live Action Role-Playing (LARP) games and similar experiences are becoming a popular game genre. Here, we discuss how artificial intelligence techniques, particularly those commonly used in AI for Games, could be applied to LARP. We discuss the specific properties of LARP that make it a surprisingly suitable application field, and provide a brief overview of some existing approaches. We then outline several directions where utilizing AI seems beneficial, by both making LARPs easier to organize, and by enhaning the player experience with elements not possible without AI. 
\end{abstract}

\begin{IEEEkeywords}
LARP, Live Action Role-Playing, Role-Playing, Game Design, Artificial Intelligence
\end{IEEEkeywords}

\section{Introduction}

How could an artificial intelligence (AI) help a pretend paladin hunt an orc through a forest? The research field of AI in Games, which should technically be able to answer this question, separates into two sub-fields. One dealing with how to make AIs that can play games to win, the other sub-field asking how AI can enhance the game experience. For an AI to play a live action role-playing (LARP) game well seems to be a somewhat far-fetched endeavor. It is not unimaginable that some day we will have a fully embodied AI that convincingly plays make-believe in a shared imagined world, navigating both the physical and the interpersonal challenges, while figuring out how to actually ``win'' in an open-ended, game-like interaction. For now, we are more interested in AI applications to enhance the game experience in LARP. Naively, AI might seem like a poor fit for a game genre that is often associated with a deliberate lack of modern technology. But there are already early attempts to integrate modern technology into LARP \cite{segura2017design,Segura2018, Dagan2019}.
And AI in games research in the past has focused on game design, believable characters, world building, story telling, automatic game balancing and player modeling \cite{YannakakisT15} - which all sound relevant for LARP. In this paper, we argue that there is a role for AI in LARP -- especially when focusing on those AI technologies that have already been successfully applied to other game genres.




First, we will give a short overview about what LARP is, and related it to similar game forms, such at tabletop role-play (TRP). We discuss specific properties, that make LARP both a suitable application for AI research, while also providing unique and new challenges. We specifically talk about the decomposition of the different functions that are usually all performed by a single game master in TRP. We then take a brief look at existing applications, both in LARP-like domains, and those that may easily be adapted. Finally, we put forward suggestions on how AI can both address existing challenges and further enhance game-play beyond what is possible. This part has both some more generic suggestions, as well as concrete illustrative examples. The overall goal of this paper is to point out possible avenues of AI in Games research in the underutilized domain of LARP.

\section{What is Live Action Role Play}


There are a range of LARP definitions which vary dependent on what tradition they are from. Salen et.al.~\cite{salen2004rules} see LARP as a descendant of TRP that takes place in a real physical space, where people act out their characters and their actions. Particularly in the US and the UK many early LARPs~\cite{harviainen2018live} were embodied games situated in fictional worlds based on either ``Dungeons and Dragons\cite{gygax1974dungeons,peterson2012playing}'' or ``Vampire the Masquerade\cite{rein1992vampire}''. LARP also has similarities with Reenactment, but differs in that the outcome of events, such as famous battles, are less predetermined, and more dependent on player's actions. 

One might also explain LARP as an improvisational theater play in which one is playing a character, knows about the other characters and the world, but does not have a script to follow \cite{stark2012leaving}. This definition resonates with another LARP tradition, referred to as theater LARP, which generally focusses on character interaction and relationships, and is more lightweight on the rules. There are similarities to experimental theater, but players are active participants, play characters, and rely on a conflict resolution mechanic to determine what happens if two participants have conflicting ideas on how to progress the story. A classic example here is the murder mystery party, particularly a party in the later style where participants are assigned characters and become supporting players in each other's experience. 

There are many other traditions and corresponding descriptions of what LARP is, see~\cite{harviainen2018live}. Central are usually a physical embodiment of at least some players as their character. Beyond that, there is a lot of diversity in existing LARP events nowadays. In the following section we discuss some properties of LARP, with a particular focus on those issues we think are relevant for AI facilitation in these games.













\subsection{Decomposition}

\begin{figure}[h]
  \centering
  \includegraphics[width=\linewidth]{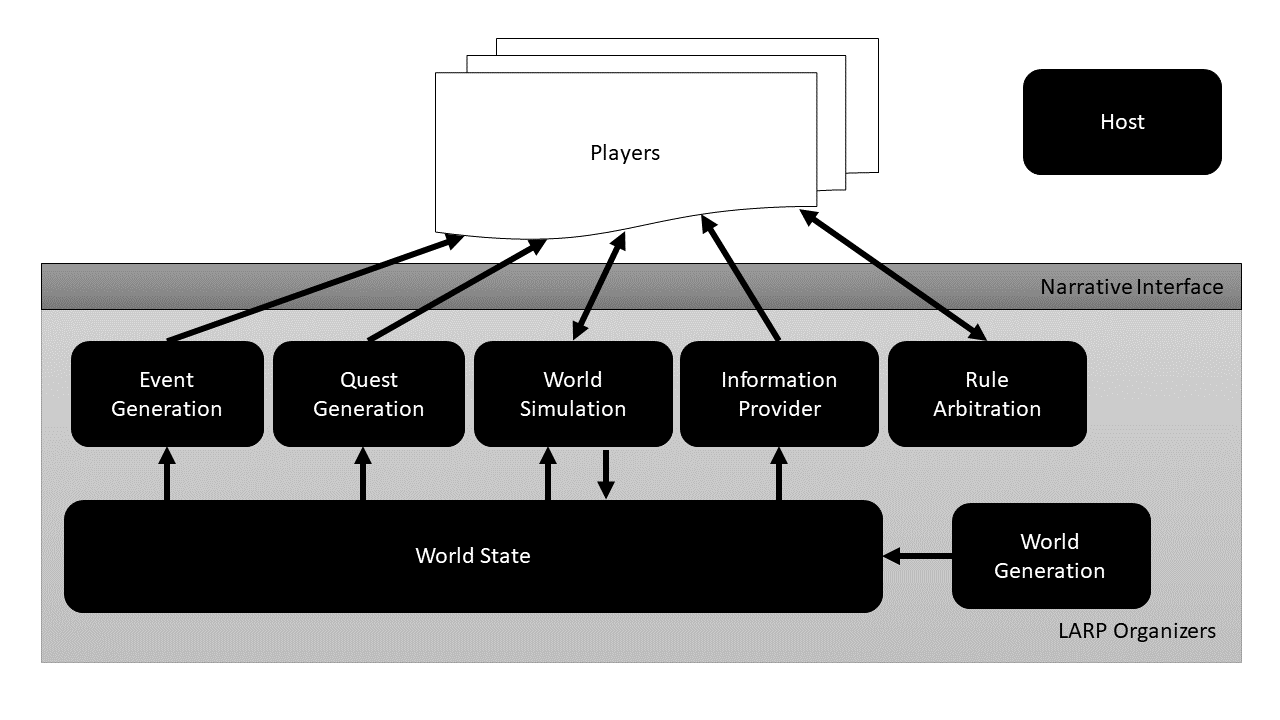}
  \caption{Decomposition of different functions\,/\,roles for LARP organizers. Arrows denote flow of information. In tabletop role play all functions in the grey box are usually performed by the game master, a single person, and all interactions are mediated by a single, narrative interface. The interface in LARP is between each function and the players, and can take different forms.  }
  \label{fig:decompose}
\end{figure}

A relevant distinction between table-top and live role-play is the decomposition and realization of the different functions of the game master. In TRP a single game master acts as a storyteller, quest provider, information provider, arbiter of rules, world simulator, and often also host. Most of those functions operate on the fictional world that resides in the GMs mind. Players interface with this world verbally by stating their actions and the results are relayed to them via speech. In LARP, particularly in larger ones, these different functions are performed by different crew members, other players, and the actual physical world. It is typical that the different members in a LARP organization team have different roles and responsibilities. While some crew members might be in charge of the game's plot, other might just be there to ``monster'' \cite{mitchell2019volunteers}, i.e. play opponents and non-player characters, without having an understanding of the overall game or plot.  This decomposition introduces challenges, but is also an opportunity to solve different aspects of the AI game master problem separately. Fig.~\ref{fig:decompose} provides a diagram of the functions we now describe in detail. We should note that this is largely oriented at big, entertainment focused, UK-style LARPS, such as Empire, etc. Other LARP traditions, such as Nordic LARP \cite{stenros2010nordic}, do have similar roles, but their functions or limitations might be slightly different. 

\subsubsection{World State} LARPs operate in a game world that is at least partly fictitious. Facts like who is dead, or who rules a given fictional nation only exist in this fictional game state. Players can and do create their own collaborative fiction, which unlike in TRP is hard to monitor for the organizers, but still part of the fictional world state.

\subsubsection{World Simulation} Roleplay can be described as a conversation between players and referees - where one side says, what do you do? and the other asks, what happens? When players act in the world, the effect of their actions has to be determined. Unlike TRP, The simulations are  a collaborative effort between the players and the organiserss of the LARP, Updating the world state is time-consuming and requires significant coordination and skill.

\subsubsection{Event Generation / Storytelling}

While there are LARP events that are purely driven by player actions, many feature a plot presented to the players in the form of events. In a TRP this might be realized by the game master prompting groups of players towards certain outcomes. The process cannot be trivially individualised, which limits the ability of the GMs to adhere to different perspectives.

\subsubsection{Quest Generation} Another way to get players to act is to give them quests. While this is common in TRP, and is often woven into the play around the table, it is less common in LARP. By their very natures, LARPs are much more freeform, with players expected to act intrinsically. Story paths devised by players can, however, pose a challenge to the coherence of the game, as they might not be aligned with the overall scope.  

\subsubsection{Information Provider} Another common role for non-player characters is to provide information about the world to the player. Since some of the game happens in a fictitious world that cannot be directly observed by the player, it is important to inform them about events or relevant facts. This in turn ends up requiring constant player briefing, which again creates limits the ability of the help players ``experienc'' to tell a story effectively.  

\subsubsection{Rules Arbitration} Most LARP events rely on an honor system in which player self-police their compliance with the game rules. For example, it is common that players have life points, and count themselves how many they lose in any given fight, determining themselves when they die. Nevertheless, it requires from players to constantly monitor and report their own state and the state of other players. As we will see in the related work, there are also technical solutions to track such things. 

\subsubsection{Hosting} In addition to managing the game's fiction, a LARP organizer is usually also responsible for hosting the event. This might include ensuring that there are adequate sleeping, food and hygiene arrangements, and that everyone is feeling safe and comfortable. This creates obligations to ensure a space that is free of dangers, harassment, etc. As with other games, it should be possible to withdraw from the game at any point without any negative real world consequences.

\subsection{Coherence}
\label{sec:coherence}

The size of LARP events usually forces organizers to split the previously mentioned roles across different people - which are often physically separated. So, while in an ideal world every meaningful player action would update the world state, and this update would be communicated to all relevant parties, that level of responsiveness is often not feasible. Additionally, non-player characters might lack certain pieces of information. Information might also be misremembered by players. Because there is no objective external reality against which this information can be verified, wrong information propagates more within the LARP than it would in real life. For example, a player who hears a rumour that the sky is red cannot simply glance up and disprove it. Taking into account that in a LARP there is locality, that is, the acting persons are usually not in the same location, the decoherence is to a certain extent also necessary in order to properly represent the game world. E.g., if an important new information is created in one location of the LARP game world, it may travel by means of gossip to more distant participants, but it would be unrealistic to distribute it instantly to all players, even if that would be technically possible.

Communication technology can address many of these issues, and some LARPs do rely on radios, wifi and databases to combat these problem. AI can cam into play and help monitor certain aspects of the player's LARP ``life'', so as to have a coherent picture delivered to every single player. 

\subsection{Physicality} One defining characteristic of LARP, in contrast to TRP, is that at least some of its play is carried out in the physical world. This makes certain things easier - actions are sometimes easier to perform than describe, and certain elements of the world are more easily ``simulated'' in the real world than in someone's mind. For example, a physical sword fight in a LARP is usually quicker than a similar, simulated fight in Dungeons and Dragons \cite{gygax1974dungeons}.  The challenge for AIs here is to operate on potentially faulty, incoherent and incomplete information, and then affect the physical world. Several challenges from the domain of human-robot and human computer interaction also arise naturally --- such as ad-hoc negotiation of collaboration or cooperation in a physical space.


\subsection{Scalability}
 A lot of bottlenecks deal with the question of how the different parts of the organization exchange information and keep the game coherent. Scaling up a LARP often means adding more people for those roles, or splitting the roles among more people, and hence increases the challenge of keeping everyone in sync. In addition, scalability is an issue for generating quests or meaningful game content for more and more players. LARPs often feature a similar amount of plot regardless of size - resulting in a sparsity of things to do the more players there are around. This is often related to the problem of content creation. Things like background stories and world information usually have to be written, or at least reviewed by the organizers, and care has to be taken that all those stories are consistent with each other.  

\subsection{Immersion} Another characteristic of LARP is immersion, i.e. creating the feeling of genuinely being there or experiencing what you are playing. A popular goal for LARP is the so called 360 immersion \cite{koljonen2007eye}, the idea that everything around you conforms to the scenario - that what you see and feel would be similar to what your character would see and feel. In a fantasy LARP this could mean banning drinking cans, making sure costumed look, or even are, authentic, and that there are no visible modern buildings in sight. AI devices can help by making this experience feel more authentic (e.g. spirits bound in trees can be virtual agents --- more on this later). 


\subsection{Robustness}

In a lot of LARPs, particularly those with less commercial routes \cite{harviainen2018live}, there is an understanding among players that they are not just consumers of an experience, but are actively helping to create the same experience for others. Player are usually willing to overlook minor problems and help to make LARPs work. This comes both in the form of being willing to adapt and interpret inconsistent clues, and in a willingness to improvise to fill the gaps. This gives LARPs a certain degree of robustness, in contrast to i.e. computer RPGs, where an AI-player or AI-story must work, otherwise the game will crash. In LARP players might just fix minor problems.








\section{Related Work}
\label{sec:related}

We will now look at a selection of existing LARP and LARP like experiences. The point we try to make with these example is as follows: There are already examples of technology facilitation in LARP. Technology can make LARP easier to organize and run, and enable interaction not possible without said technology. This provides an interface, and even existing world representations that common AI in Games approaches could built upon.

\subsection{Existing LARPs and related projects}
\label{sec:existing}
\subsubsection{Wing and a Prayer}
is a LARP based on the experience of the Women’s Auxiliary Air Force (or WAAF) during World War 2, supporting the Royal Air Force \cite{ianthomasWaPStructure}. Female players interact with realistic radar set-up and data, and the outcome of simulated battles depends on the quality of instructions they give to pilots, played by male players. In this LARP, a computer simulation provides a more mimetic experience of wartime conditions than would otherwise have been possible. It is an existing example of how the complexity of world simulation can be offloaded on a computer, and how a system can be designed that allows players to interact with such a digital representation, without losing their immersion. This use of technology to mimetically portray information-giving devices mirrors one found in many escape rooms \cite{nicholson2015peeking}. Escape rooms are experiences designed for a small number of players (typically 4-8 at a time) over a shorter period of time (a 60 minute time limit is common) and with a heavy reliance on props to enforce all the rules of the game. In that context, it is common to find encryption devices, hackable computers, and similar real or simulated technology.

\subsubsection{Bad News} is a theatrical game where a single player interacts with a an actor, who portrays different characters from a procedurally generated American town, based on instructions from a computer \cite{Samuel2016}. While the physical interaction is limited to embodied conversation, this project illustrates how procedural generation can be used to generate a world state, and how an interface between a digital representation and an actor can be used to produce a range of characters that are consistent with the world model.



\subsubsection{Empire} is a UK based fantasy LARP with hundreds of players\footnote{\url{https://www.profounddecisions.co.uk/empire-wiki/Main_Page}}. Interesting to us is the fact that Empire maintains a database that stores a range of information for every player character. Between events, players can use an online interface to decide how they want to use their assets. They can send their warrior bands to support certain military conflicts, or send their ships to trade with specific foreign nations. Before each actual physical event, those inputs are processed, and as a results players are given information and resources for the actual event. Player might also perform actions during the event that affect this database, such as performing a ritual enhancing the yield of their farm, or using a cleric to inscribe a word on their soul \footnote{\url{https://www.profounddecisions.co.uk/empire-wiki/Testimony}}. Empire is typical example of how an data back-end can be used to enhance gameplay, and how it can be interfaced with the players, both during play and in between sessions. 




\subsection{Technology}

Segura et.al.\cite{segura2017design} provide an overview of existing technology use in LARP, and propose a preliminary taxonomy. Existing technology has been used to simulate aspects of the world, tracking players, communication, etc. They also report on a strong focus in regards to the aesthetic experience. There are also several wearables \cite{Dagan2019,Segura2018,vanhee2018firefly} that were specifically designed for LARPs. They are all designed to be worn during play, and provide a range of game-play affordances. They allow for a display of character information, such as health or character affiliation, allowing the player to ``see'' a virtual fact that would be visible in the fiction - such as a character being hurt. These wearable also allow for a range of social affordances, such as healing a character from an ``power surge'' by touching their wearable. There is also a ``Technomancer Hoodie'' which is equipped with motion sensors that allows the wearer to ``cast'' a series of spells by making appropriate movements, which are then simulated with sounds and colored lights from the hoodie. There are also examples of apps \cite{segura2017design}, usually run on smart phones or similar devices, that are used to support LARP experiences. 
In general, there are now quite a range of technologies that support and enhance the LARP experience, and several of them offer the opportunity to interface the player with an AI system. In this paper we want to particularly focus on the use of artificial intelligence to further enhance these approaches.



\section{Possible AI applications}

In this section we want to present some sketched out examples of how existing, or conceivable AI approaches, particularly those from the AI in Games domain \cite{YannakakisT15,yannakakis2018artificial}, could be applied to LARP. The goal is to both make organizing LARPs easier, by overcoming the previously outlined challenges, and to enhance gameplay in a way not possible without AI. We will make some assumptions about existing tech to facilitate the deployment of AI, and will try to relate this back to existing technology or prototypes. We will also try to provide examples in both a generic form, talking about a range of possibilities, and then specify them, with more concrete illustrative examples. 


\subsection{Conversational Agents}

There is currently tremendous interest  from the AI community towards building conversational agents - commonly known as chatbots. The technology is not mature yet to the point where general conversations can be had. There is a debate raging currently in the AI community as to whether more data or better algorithms are needed, but within a closed domain it seems reasonable results are achievable~\cite{gao2019neural}. But how could these chatbots be used in LARP?

One common interaction between players and ref is asking for rules clarification, such as ``Can I dodge epic damage?'' This is a role that could be filled by a straightforward question-answering system (e.g. see \cite{lan2019albert} for a modern take). As this is an ``out of game'' interaction, such a system could be deployed on a smart phone, or similar device, and there would be no need to simulate a character for that AI. According to our domain experts, this would already free up a large chunk of the time spend by the organization team.

Taking this a step further would be a conversational system that could provide ``in game'' information, such as ``Who is the current ruler of this place?'' In a most simple case, this could also just be a digital device with an question-answering system trained on a fixed text corpus, detailing the background of the world. But there are opportunities for improvement here. First, if a digital representation of a world model exists, similar to the data back-end for the game Empire, the system could provide access to real time information as it changes over the course of the game. This could then also be subject to restrictions, so that certain information is not accessible to the players. This has several advantages. While human NPCs can and often do perform the role of lore-giver, that task can become repetitive, and there is the risk of human error, with the result that different groups of players get different information from the lore-provider. There is also less danger of the AI improvising new world facts.

Secondly, one could take care to embed the conversational agent as an actual part of the game world. For example, the digital device could be hidden in a prop that makes it look like a talking magical book, to increase the immersion. To further the immersion the conversational AI could be imbued with character traits that manifest in the way it speaks. Taking this even further would be to employ techniques that give an appearance of agency. Interactions with the AI could lead to changes in its mood - and the forms of interaction available might depend on past interaction. For example, the magical book might also tell certain secrets to people that where nice to it. Having the AI character adapt to past interaction could also help with a differentiation between several, similar AI objects, who could have different relationships, or even different amounts of knowledge. Not all AIs might know the same world information, and they could even learn new information over time. 

Chatbot-like digital games such as Event[0] (Ocelot Society), Don't Make Love (Maggese), and projects by LabLabLab\footnote{\url{https://www.lablablab.net/}} demonstrate some of the possibilities in this space, and some VR games using Alexa have pioneered the combination of chatbot effects with speech-to-text, as for instance in Starship Commander, where the player directs the starship using voice instructions \cite{KuelzStarship}. A similar command system could be deployed in the context of LARP, with the AI receiving voice commands and narrating information about what is happening next in some part of the game world. There is also increasing interest in the concept of "virtual humans" — persistent non-player characters who might interact with a player via a combination of games, VR experiences, and social media. The technology developed to support such characters for marketing, theme park, and entertainment applications might also be suitable for deployment in LARPs.

Providing real influence to conversational AI might also be an interesting and surprising way to increase the players' influence on the world. One of our interview practitioners suggested including AI gods, in the form of totems. Gods in LARP systems are often associated with specific ideals and rules, and as such could have distinct and clearly defined character traits and opinions. The could be embodied in a holy object, which would allow the player to talk to their respective deity. An additional boon here would be the fact that AIs could store their past interaction with a specific player, and reference back to those, even with month of time passing in between. 

While unlimited, open-ended understanding of input is beyond the reach of current systems, though there are currently strides being made towards this direction~\cite{adiwardana2020towards}, the fictional context of prayer or divinatory question-asking might allow the game-master to teach the players a reliable set of conventions for interacting with the conversational agent. Ritualised language use is already deployed in some contexts in table-top RPGs, such as Ben Lehman's Polaris: Chivalric Tragedy at Utmost North.

Initially, an AI conversationalist might just be seen as a way to provide information to the player, or allow for some fun role-play opportunity, but it could later be revealed that these interactions and conversations could have actual consequences. A god that learned bad information from one player about another in conversation might decide to punish them, triggering some game mechanical consequences. Or certain deities might even change their nature, based on player interaction with them. In either case, the fact that those conversational AI could be either directly communicating with the stored world state, or at least accurately store all their conversation for review between two events, would allow organizers to overcome some of the bottleneck issues with integrating player information back into the game.

\subsection{Embodied AI agents}

There are also opportunities to having the previously mentioned AI agents physically embodied in the world. Let's revisit the earlier example of a smart phone stored in a talking magical book. It could use its GPS sensor to determine its current location, and then trigger certain interactions when it is carried into a certain area. It might for example say that a certain area has a lot of magical energy floating around, or that a lot of people died in a certain space. Precedent for such a system exists for instance in Nico Czaja's work \cite{nicoczaja} with xm:lab, creating phone-based narrators who tell an interactive story while the participant wanders through a space of real-world historical significance.

Similar interaction could be possible with the previously described wearables - which could take different roles in the fiction. There are already examples of tech that mimics the role of pip boys form the Fallout Game series, a fictional wearable that informs the player about their stats, and warns them when they are entering a radioactive area. Adding conversational AI to this might turn this into a game companion that rides around on your shoulder. The locative aspect could then also be used to trigger salient character based interaction, similar to the approach used in the computer game ``Heavens Vault''\footnote{\url{https://www.gamasutra.com/blogs/JonIngold/20180822/325018/Ideas_and_dynamic_conversation_in_Heavens_Vault.php}}. There the fictional companion robot selects dialogues from a range of sentences, based on both previous interactions, locations and elements that are currently present in the scene.





\subsection{Drama Management: AI director}

How to identify the time and circumstance for a pre-defined story beat? Narrative designers of conventional video games often use a system of storylets or quality-based narrative, in which story events are triggered whenever some pre-condition state is reached in the game world. They write individual moments that they want the player to experience at some point, and then allow the system to select the point when those moments are best presented during a particular player's playthrough. An academic survey of the uses and applications may be found in \cite{kreminski2018storylet}, and an overview intended for users in the video game industry can be found at \cite{horneman2017}.

LARP creators have written about building LARPs with similar gameplay beats in mind. Ian Thomas has written about starting with "moments" in his design for both ``All for One'', a Musketeer-themed LARP \cite{ianthomasAllforOne} and ``God Rest Ye Merry'', a Christmas ghost story set in the 1950s \cite{ianthomasStuntToStory}. In the case of existing LARPs, it typically falls to human GMs to determine when the moment has arrived to deliver a story beat, and there is little room for last-minute customization. An AI system able to track key elements of world state, however, would be able to select when to activate particular storylets, and potentially use grounding techniques similar to those used in video games to fill in elements of the delivery, customizing the story moment to the exact parameters that allowed it to be fired off.


During play, the AI might also detect players who appeared to be inactive or who hadn't recently made any game play discoveries. It might then trigger events to re-engage those players, to send information or NPCs to their locations, or to move combat in their direction.

An AI drama manager that also had profile data about player preferences — whether calculated or provided to the system by players themselves — could determine what types of re-engagement should be employed for these specific players, and what type of pacing they might prefer. An adjustable system with an awareness of player profiles might help in making LARP more accessible to players with a wide range of ability, interest, ans experience levels.

Work towards modeling players and their preferred storytelling experience has been done by \cite{ThueBulitko}, and towards the problem of storytelling for specific types of player \cite{GervasPeinado}, but primarily in the context of video games or tabletop roleplaying rather than in the context of LARP.







\subsection{AI Content Generation}

Before a LARP is first run, depending on the setting, there is usually a phase where the organizers create a world and setting. This can involve writing fictional history, developing a fictional cosmology and theology, and designing existing characters and their relationships. There usually is no great need for AI support here, but systems like ``Bad News'' \cite{Samuel2016} or the Legends Mode of Dwarf Fortress \cite{DFwiki} demonstrate that it is possible to create a setting and web of relationship based on certain historical settings. Other techniques for AI story generation are surveyed in \cite{Gervas09,kybartas2016survey}, and a more recent breakdown of the key challenges to be solved in this space can be found in \cite{Gervas19}.

There are two ways in which procedural content generation could help human designers. In a mixed initiative co-creative approach an AI system could produce fictional history or relationships, and a human designer could then select and refine. The system could also be used to generate inspiration of ideas. 
On a more practical level an AI system could also provide more complexity after the rough brush strokes have been filled in by a human designer. This might be useful to engage players who want to engage in a more scholarly play style. For example, imagine the human designers have created a rough world design, with major historical events, places and characters being defined. An AI system could then fill in the gaps to create smaller places and additional characters. This result could then, for instance, be piped into a narrative generator that creates travel diaries of a minor historical character to these places \cite{ShortParrigues}. A human designer might then hide a few connected bits of information that could be combined to gain some important insight into something related to the event overarching plot. The end result could be a full book containing a range of stories about the game world, which could also afford ``academic research'' game-play, where player would study the book in detail to hunt for those bit of highly relevant information. Story generation techniques with an awareness of level-of-detail might prove valuable in this context~\cite{Flores2017LevelOD}.

Once a system like this was set up, it would be easily scaleable - in theory one could provide a whole library worth of research that is both related to the world, and allows for relevant research to be played out. A human designer usually cannot provide this amount of content due to time and labor constraints. In general, this could help to alleviate the problem of how to provide adaptive resolution in a physical setting. In a TRP a player might walk into a library and grab a random book, open it and start reading. The game master can then, on the fly, come up additional content for that book. But building a library for player to explore in the physical world is much harder, as it would require to produce that kind of complexity beforehand. 

A similar issue to this is puzzle design. Procedural design of puzzles for point and click adventures are explored in \cite{FernandezVara}. A more physically-grounded variation on this idea might also be possible to deploy in a LARP environment.


\subsection{AI Story Hooks}

Another opportunity for AI content generation is to produce story hooks or quests for players - providing for more micro-questing in larger LARPs. As previously discussed, there is not necessarily a reason to evaluate the success or failure of a quest, already providing a goal could lead to the desired outcome, more interaction and role play. 
There are already existing approaches to automatic quest generation, usually looking at existing NPCs, their role in the world ontology, and their desires \cite{Kybartas2014,lee2012dynamic,ashmore2007quest}. These tools could be easily modified to provide a range of quests to players, if a data based representation of the world and the players exists. These quests could then be offered in between events to players, as part of their event briefing package, or even prior to the event in digital form. This would allow the player to accept and reject certain type of quests - which could enable player modeling and more personalized quest generation. A player might, for example, not be interested in performing any illegal activity, as it clashes with their character concept. 

Another way to approach this problem would be the generation of story hooks rather than quests. One way to provide narratives for computer games is to create a range of characters and associated conditions and then run a simulation to see what happens \cite{meehan1977tale,kybartas2016survey}. This approach has been quite popular for creating murder mystery plots \cite{stockdale2016cluegen,jaschek2019mysterious,mohr2018eliminating}, where agents are simulated until a murder happens, that can then be reconstructed by the player. We might do something similar in LARP, to create investigative mysteries, possibly with game relevant information, such as involving characters the players previously encountered. But there is also another application. We could model the player characters as virtual agents, and then try to assign them goals and resources, and simulate what would happen. After running multiple simulations we could then select a set of starting conditions that looked like it created an interesting story. Here we would not explicitly create a narrative, but rather the conditions that could potentially lead to one. For example, one character might be given the goal of finding a suitable partner for one of his siblings sons - while to other players might be given corresponding quest of finding a suitable match for a daughter. These story hooks could be customized in several ways to fit the given character, and adapt to existing family situations, preference for certain kinds of roleplay, character traits, etc. The simulation of what could happen could also help to design story hooks with a certain degree of redundancy and robustness. Overall, this could provide additional hooks for player to inspire their roleplay, and could fill in some of the sparsity of plot in larger LARP events. 





\subsection{AI Aided World Simulation}

A big element and defining characteristic of LARP is how the player interacts with the real and the game world. Interaction with the fictional world usually requires a referee - and as such provide a bottleneck that designers try to minimize. AI could alleviate this by providing part of this complex, fictitious world simulation. One example here is a spaceship larp, where players interaction with various ship systems, such a engineering, navigation, piloting, resource control could be tracked, integrated and fed back to them in real time. AI simulators are able to integrate large amount of data and their effects, and even provide this preprocessed information to the organizers to reflect the collective player agency.

Another example is the use of AIs to evaluate player rituals. In UK LARP ritual magic is often performance-like and open ended, requiring a referee to determine how the intent of the players is translated into a game world effect. This evaluation is time consuming - to illustrate, a system like ``Guild of Darkness'' has one crew member who's full time responsibility is to evaluate rituals\footnote{\url{https://sites.google.com/site/guildofdarknesswiki/magic}}. Evaluation is based on performance, intent and in part dependent on some (hidden) world rules - which the referee has to consistently apply, so player can learn more about the world.


Part of this process could be automated by a neural network looking at the movements performed in the ritual. A system like Wekinator \cite{fiebrink2010wekinator} can be trained to take a sensory input, such as an image, and assign a sound or output value. This could be used to design a system where certain moods or elements need to be triggered at given stages (based on the desired effect), and the quality of the ritual would depend on how well these queues were hit. This would foreground the AI system more, and require some game design around the AI system, but has some advantages. Performing magic in a ritual is usually an attempt to understand an please an arcane and unknown system. Having this simulated by an AI would give the system a consistency that a human cannot provide, and would allow for a nearly arbitrary scaling up of the underlying complexity. So there could be a process of player getting successively better and better at understanding and triggering this system - allowing for role-play of actual skill development. It might even allow for differentiation, where different player are better at different types of rituals. 

This idea could be even taken further - by designing a magic system directly around it. Imagine a system where there is specific magic site where player can perform movements in front of a camera, or other sensors, camouflaged as a mythical artifact. Specific inputs could be assigned to words, and as the player perform certain movement they could discover how to trigger these words. Part of the game-play could then be the discovery and refinement of the inputs the create these words. There should be some feedback, such as the mythical artifact uttering the triggered words, or even some indication once player get closer. Ultimately, the idea would be that those words have to be formed into sentences, which would then allow the players to talk to some entity. This could then be the organizers, or even one of the previously discussed god AI NPCs.  
\subsection{Super LARP and Information Spread}
\label{sec:superlarp}
Once AI support and AI characters allow for better information transfer, information integration and game cohesion, more ambitious projects, such as coordinated over different locations or times. Next to the scalability issue, the most important problem for such an endeavor seems to be the organization of the information flow, which ties in with the consistency issue. Even for a large LARP as run nowadays, there is already a physical distribution of actors and information and this could be taken into account for by applying epidemic protocols for gradual information flow in distributed databases as suggested in \cite{Demers1987}. This type of information distribution is also called gossip communication~\cite{Jelasity2011} and is models very well how news spread in a distributed human society. Once such protocols would be established, different locations could play out organizations or events that mostly interact through information sharing, negotiation, or how they affect a joint world.

\section{Conclusion}

In summary, we believe that LARP is a domain well suited for the application of AI and AI and Games techniques. The listed, existing approaches demonstrate this, and the speculative examples show a range of relatively straight-forward extensions of existing AI and Games techniques, so they would be suitable for LARP. Doing so could overcome several existing challenges for LARP organizers, such as scalabiltiy and content generation issues. It could also provide for new forms of play that would not be possible without AI. LARP also provides an interesting test-bed for AI applications, particularly those that want to explore the interface between humans and AI, or how AI can interact with the physical world. Here the robustness of LARP, caused by the willingness of the participants to correct errors on the fly, could provide valuable for researchers. In general, AI in LARP research offers several unexplored opportunities, both to enhance the experience of players, and to explore the limitations and challenges of AI.

\section{Acknowledgments}
This paper is in large part based on a discussion group at the Dagstuhl Seminar 19511, ``Artificial and Computational Intelligence in Games: Revolutions in Computational Game AI''. We are also grateful to Jonathan Henderson and Ben Crossley, both of them UK based LARP organizers, who we interviewed to sanity check out assertions in this paper. 


\bibliographystyle{IEEEtran}
\bibliography{sample-base}
\end{document}